# Interpretable Machine Learning for Air Pollution and Respiratory Health Prediction: A Socioeconomic Subgroup Analysis

Maede Azani Hassan Abadi, Shouyi Wang

Department of Industrial, Manufacturing, & Systems Engineering, The University of Texas at Arlington

Corresponding author: mxa5459@mavs.uta.edu

## Abstract

Air pollution and climate-related stressors are increasingly important concerns for respiratory health, especially in settings with unequal environmental exposure and healthcare capacity. This study evaluates an interpretable machine learning framework for predicting respiratory disease rates and air-quality status using structured country-level weekly data. The analysis used a publicly available climate-health dataset containing 14,100 observations from 25 countries between 2015 and 2025. Two supervised learning tasks were considered: regression of respiratory disease rate per 100,000 population and binary classification of air-quality status. Nine regression models and nine classification models were compared using nested cross-validation. Model interpretation was conducted using SHAP values, and subgroup analysis was performed across income levels and geographic regions. The results showed that PM2.5 concentration was the dominant predictor of respiratory disease rate, with linear and regularized linear models achieving the strongest regression performance. For air-quality classification, models achieved high balanced accuracy when PM2.5 was included, but performance decreased substantially when PM2.5 was removed, indicating strong dependence on pollutant-related information. SHAP analysis showed that, without PM2.5, socioeconomic and meteorological variables such as GDP per capita, precipitation, and healthcare access became more influential. Subgroup analysis showed similar aggregate regression error across income groups, but PM2.5 contributed more strongly to predictions in lower-middle-income countries. These results show that model accuracy alone is not sufficient for climate-health prediction. Interpretable models can help identify dominant pollution-related signals, test whether results depend on key pollutant variables, and show whether prediction patterns differ across socioeconomic groups.

## Highlights

- An interpretable machine learning workflow was evaluated using a public dataset of 14,100 weekly country-level records.
- Observation-level nested cross-validation compared nine regression and nine classification models; the reported scores represent within-dataset performance.
- PM2.5 was the dominant predictive signal for respiratory disease rate, but this model-based association is not a causal or epidemiological estimate.
- Removing PM2.5 from the AQI-derived classification task substantially reduced balanced accuracy, revealing strong target-proxy dependence.
- Subgroup analyses found similar aggregate errors but different feature-attribution patterns, which should be validated using independently traceable data and grouped or temporal validation.

# 1. Introduction

Air pollution and climate-related environmental stressors are major contributors to population-level respiratory health risks. Fine particulate matter, particularly PM2.5, has been widely associated with respiratory and cardiovascular morbidity because it can penetrate deeply into the respiratory system and contribute to inflammation, disease exacerbation, and increased healthcare burden (Ku et al. 2022; Cappelli et al. 2024; Atzeni et al. 2025). At the same time, climate-related factors such as temperature anomalies, heat waves, and precipitation patterns may influence respiratory outcomes directly or indirectly by affecting pollutant formation, exposure patterns, healthcare access, and population vulnerability (Ku et al. 2022; Cabral-Miranda et al. 2025; Im et al. 2025). Understanding these relationships is especially important for public health surveillance and environmental policy, particularly in settings where exposure and health risks are unevenly distributed across socioeconomic groups.

Machine learning provides a useful framework for studying climate-health relationships because it can integrate heterogeneous predictors, including pollution measurements, climate indicators, socioeconomic conditions, geographic characteristics, and temporal variables. Compared with traditional statistical models, machine learning methods can evaluate a wider range of functional forms and interactions, making them attractive for environmental health prediction tasks. More broadly, learning-based methods have been increasingly used to capture complex structures and support prediction and decision-making problems (Ardali et al. 2026). In environmental health research, recent studies have applied machine learning to air-quality forecasting using methods such as gradient boosting, XGBoost, and Gaussian process regression (Han et al. 2023; Saminathan and Malathy 2023), to respiratory disease prediction using random forests and neural networks (Ku et al. 2022; Cabral-Miranda et al. 2025; Yehuala et al. 2024; Cappelli et al. 2024), and to health surveillance and environmental risk assessment using XGBoost and interpretable models (Telesca and Rondinone 2025; Im et al. 2025). However, many applied studies prioritize overall predictive accuracy over interpretability (Houdou et al. 2024), give insufficient attention to validation design and the risk of overly optimistic performance estimates (Raschka 2020), and rarely examine prediction error across socioeconomic or geographic subgroups (Mehrabi et al. 2022; Mhasawade et al. 2021).

Several methodological issues are particularly important in climate-health machine learning. First, model evaluation must avoid overly optimistic performance estimates. When hyperparameter tuning and model evaluation are performed on overlapping data, reported performance can be biased upward. Nested cross-validation provides a more rigorous evaluation structure by separating model selection from final performance estimation (Raschka 2020). Second, predictive models should be interpretable, especially in public health applications where results may inform surveillance, risk communication, or resource allocation. SHAP-based interpretation provides a consistent way to estimate the contribution of each feature to model predictions and has become widely used in environmental and health-related machine learning studies (Houdou et al. 2024; Ku et al. 2022; Telesca and Rondinone 2025; Im et al. 2025). Third, classification tasks based on environmental indices require careful sensitivity analysis, because pollutant variables may be closely related to the construction of air-quality labels. Without examining this relationship, classification performance may appear stronger than it would be under a more conservative feature setting. Finally, aggregate model performance may conceal differences across socioeconomic or geographic groups. In environmental health contexts, subgroup evaluation is important because vulnerable populations may experience different exposure-response patterns and different prediction errors (Mehrabi et al. 2022; Rajkomar et al. 2018; Mhasawade et al. 2021; Rojas et al. 2022; Schuch et al. 2023).

This study addresses these issues through an interpretable machine learning analysis using the publicly available Global Climate-Health Impact Tracker dataset. The dataset contains 14,100 weekly country-level observations from 25 countries between 2015 and 2025. Although the dataset documentation identifies broad source categories, it does not provide complete variable-level provenance or a fully reproducible procedure describing how the sources were processed, harmonized, and combined. The study is therefore framed as a methodological demonstration of model evaluation and interpretation rather than as an epidemiological assessment. Two supervised learning tasks are evaluated. The first predicts respiratory disease rate per 100,000 population using climate, air-quality, socioeconomic, geographic, and temporal predictors. The second classifies weekly air-quality status into good or unhealthy categories. Multiple model families are compared, including linear models, tree-based models, ensemble methods, kernel-based models, neural networks, and instance-based learners. Observation-level nested cross-validation is used to separate hyperparameter selection from outer-fold performance estimation, and statistical comparisons are reported with appropriate caution regarding dependence among cross-validation folds.

Beyond predictive performance, this study applies SHAP analysis to describe the behavior of fitted models and identify the features most strongly associated with their predictions. Because PM2.5 is closely related to the AQI used to construct the classification target, an additional sensitivity analysis removes PM2.5 from the feature set. The PM2.5-included analysis is treated as a diagnostic benchmark, while the no-PM2.5 analysis provides a more conservative assessment of classification from indirect environmental and socioeconomic signals. The study also reports descriptive subgroup performance across income levels and geographic regions to examine whether aggregate results conceal differences in error patterns or model attribution.

This study contributes a transparent methodological demonstration in four ways. First, it compares multiple machine learning models under a nested cross-validation framework. Second, it uses SHAP analysis to describe which environmental and socioeconomic variables drive predictions in selected XGBoost models. Third, it evaluates the sensitivity of AQI-derived air-quality classification to the inclusion of PM2.5, highlighting the importance of checking target-proxy relationships. Fourth, it examines performance and attribution patterns across socioeconomic groups. These contributions concern model evaluation and auditing within the analyzed dataset; they do not establish causal exposure-response relationships or population-level health effects.

The remainder of this paper is organized as follows. Section 2 reviews related work on machine learning for air-quality and respiratory health prediction, interpretable machine learning, and fairness-aware evaluation in health applications. Section 3 describes the dataset, target variables, preprocessing steps, and study design. Section 4 presents the machine learning models, validation strategy, statistical comparison procedure, SHAP-based interpretation, and subgroup analysis methods. Section 5 reports the predictive performance, sensitivity analysis, interpretability results, and subgroup findings. Section 6 interprets the methodological findings and their limits. Section 7 presents the study limitations and future research directions. Section 8 concludes the paper.

## 2. Related Work

### 2.1 Machine Learning for Air Quality and Respiratory Health Prediction

Machine learning has been widely applied to environmental health problems, particularly for air-quality forecasting, pollution exposure assessment, and respiratory health prediction. Prior studies have shown that air pollutants, especially PM2.5, are strongly associated with respiratory disease occurrence, hospital visits, and broader population-level health outcomes (Ku et al. 2022; Cabral-Miranda et al. 2025; Cappelli et al. 2024). In these applications, machine learning models are useful because they can combine meteorological, pollution, socioeconomic, and temporal predictors within a single predictive framework.

Several studies have used supervised learning models to predict respiratory disease outcomes from climate and pollution-related variables. For example, Ku et al. (2022) used machine learning models to predict respiratory disease occurrence using climatic and air-pollution factors, showing that environmental indicators can provide useful predictive signals for public health surveillance. Cabral-Miranda et al. (2025) applied artificial intelligence methods to predict pediatric respiratory hospital care using clinical, pollution, and climatic variables. Yehuala et al. (2024) compared several machine learning classifiers for predicting acute respiratory tract infections and used SHAP analysis to identify important determinants. Other recent studies have applied random forests, XGBoost, support vector machines, and interpretable machine learning approaches to respiratory hospital visits, cardiovascular and respiratory outcomes, and chronic respiratory disease exacerbations (Cappelli et al. 2024; Atzeni et al. 2025). Across these studies, particulate matter and other air pollutants are frequently identified as important predictors, although model performance varies depending on data source, geographic context, target definition, and validation design.

A related body of research focuses on air-quality prediction and classification. Machine learning models such as random forests, gradient boosting, XGBoost, LightGBM, support vector machines, and neural networks have been used to forecast AQI values, predict PM2.5 concentrations, and classify air-quality conditions (Abdelmalek et al. 2025; Saminathan and Malathy 2023; Han et al. 2023) . Ensemble methods often perform strongly in these settings because they can capture nonlinear relationships among pollutant concentrations, meteorological factors, and temporal patterns. Morapedi and Obagbuwa (2023) further showed that machine learning can support PM2.5 prediction in data-sparse urban contexts, while Patel et al. (2025) reviewed machine learning and deep learning models for PM2.5 and PM10 concentration prediction. However, when air-quality labels are derived from pollutant-based indices, the relationship between input features and target construction requires careful interpretation. In such cases, high classification performance may partly reflect the inclusion of features that are strongly related to the index used to define the target. This issue motivates the sensitivity analysis conducted in the present study.

## 2.2 Interpretable Machine Learning and Model Evaluation

As machine learning models become more common in environmental and health applications, interpretability has become an important methodological requirement. In public health contexts, predictive accuracy alone is not sufficient; decision-makers also need to understand which variables are driving model predictions and whether those patterns are plausible and actionable. SHAP, introduced by S. Lundberg and Lee (2017), provides a consistent feature-attribution framework based on cooperative game theory. S. M. Lundberg et al. (2020) later developed TreeSHAP, which enables efficient computation of SHAP values for tree-based models such as random forests, gradient boosting, and XGBoost.

Recent environmental health studies have used SHAP and other interpretable machine learning tools to explain model predictions for air pollution, respiratory outcomes, heat-related illness, and other climate-sensitive health risks. Houdou et al. (2024) reviewed interpretable machine learning approaches for air pollution forecasting and highlighted the growing use of SHAP in this field. Telesca and Rondinone (2025) applied interpretable machine learning to examine interactions between cardiorespiratory diseases and meteorological-pollution sensor data. Im et al. (2025) used an XGBoost-SHAP framework to predict heat-related illness risk from meteorological data. These studies show that interpretability can help connect predictive models to domain knowledge by identifying influential pollutants, meteorological variables, and socioeconomic factors. However, SHAP is often used only as a post-hoc explanation tool after a single best-performing model is selected. Fewer studies combine interpretability with systematic model comparison, sensitivity analysis, and subgroup evaluation.

Reliable model evaluation is also essential in applied machine learning. Standard train-test splits or non-nested cross-validation can produce optimistic results when the same data are used for both hyperparameter tuning and performance estimation. Varma and Simon (2006) demonstrated that cross-validation can lead to biased error estimation when model selection is not properly separated from evaluation. Wilimitis and Walsh (2023) also emphasized practical considerations for cross-validation in healthcare model development and evaluation. Nested cross-validation addresses this issue by separating the model selection process from the final evaluation process. This is particularly important when comparing several model families with different hyperparameter spaces. Rajkomar et al. (2018) further emphasized the importance of rigorous model evaluation, algorithm selection, and statistical comparison when comparing machine learning models. In the environmental health literature, however, many studies still report only aggregate performance metrics without formal uncertainty assessment or corrected pairwise comparisons.

### 2.3 Fairness and Subgroup Evaluation in Health Machine Learning

Fairness and subgroup evaluation are increasingly recognized as necessary components of health-related machine learning. In clinical and public health settings, aggregate performance metrics can conceal systematic differences in model behavior across demographic, socioeconomic, or geographic groups. Rajkomar et al. (2018) argued that fairness evaluation is necessary for machine learning systems intended to support health equity. Mhasawade et al. (2021) discussed algorithmic fairness in public and population health and emphasized that bias can arise from data, model design, and deployment context. Rojas et al. (2022) proposed a framework for integrating equity into clinical machine learning models, showing that subgroup evaluation can reveal disparities that are not visible in overall performance metrics. Chin et al. (2023) similarly emphasized the importance of addressing algorithmic bias when machine learning systems are used in health-related decision contexts.

In environmental health applications, subgroup evaluation is particularly relevant because exposure, vulnerability, and adaptive capacity are not evenly distributed across populations. Lower-income regions may experience higher pollutant exposure, weaker healthcare access, and greater sensitivity to climate-related stressors. As a result, the same environmental predictor may have different implications across socioeconomic contexts. Although fairness-aware machine learning has been studied in healthcare more broadly, it remains less common in air-quality and climate-health prediction studies, where performance is often reported at the aggregate level only. Schuch et al. (2023), for example, showed that machine learning models can exhibit fairness disparities across socioeconomic groups in preventive health applications, reinforcing the need for subgroup-level assessment in public health modeling.

The present study builds on these three areas of prior work by combining comparative machine learning, model-attribution analysis, sensitivity analysis for air-quality classification, and descriptive subgroup evaluation across income levels and geographic regions. Rather than focusing only on the highest-performing model, the study emphasizes transparent evaluation of model behavior, feature dependence, and subgroup variation within a public analytical dataset.

## 3. Data and Study Design

This study used the Global Climate-Health Impact Tracker (2015–2025), a publicly available dataset. It contains 14,100 weekly country-level records from 25 countries between January 2015 and October 2025. Each record represents one country-week and includes environmental, socioeconomic, geographic, temporal, air-quality, and health-related variables. According to the dataset documentation, the listed source categories include World Bank climate data, the WHO Global Health Observatory, climate reanalysis models, air-quality monitoring networks, and national health statistics (Gokhale 2025). The dataset is synthetically generated to simulate realistic climate-health relationships across countries and time periods. As such, findings should be interpreted as methodological demonstrations rather than direct epidemiological estimates.

The dataset contains 30 original variables grouped into six main categories: geographic, socioeconomic, climate, air-quality, health-related, and temporal variables. A summary of the variable groups is provided in Table 1.

*Table 1- Summary of variable groups in the dataset*

| Variable group | Variables included |
|---|---|
| Geographic variables | country name, region, latitude, longitude |
| Socioeconomic variables | income level, population, GDP per capita, healthcare access index, food security index, mental health index |
| Climate indicators | temperature, temperature anomaly, precipitation, heat-wave days, drought indicator, flood indicator, extreme weather events |
| Air-quality variables | PM2.5 concentration, air quality index |
| Health-related variables | respiratory disease rate, cardiovascular mortality rate, heat-related admissions, vector disease risk score, waterborne disease incidents |
| Temporal variables | year, month, week |

Two supervised learning tasks were formulated. The first task was a regression problem in which the target variable was respiratory disease rate, measured as weekly respiratory illness rate per 100,000 population. The objective was to evaluate how well climate, air-quality, socioeconomic, geographic, and temporal predictors reproduced within-dataset variation in this target. The resulting predictions and feature attributions are model-based associations and should not be interpreted as causal effects.

The second task was a binary classification problem in which the target variable was air-quality status. The binary label was constructed from the air quality index using a threshold of 100, where observations with AQI values greater than 100 were labeled as unhealthy and observations with AQI values of 100 or lower were labeled as good. The resulting class distribution was relatively balanced, with 54.2% of observations classified as good and 45.8% classified as unhealthy.

Because the classification target was derived from AQI, the AQI variable itself was excluded from the classification feature set to avoid direct target leakage. PM2.5 was examined separately because it is closely related to AQI and strongly associated with the derived label. Two classification settings were therefore evaluated: a diagnostic setting with PM2.5 included and a more conservative setting with PM2.5 removed. The difference between these settings was used to evaluate target-proxy dependence rather than to claim independent forecasting of regulatory air quality.

The preprocessing workflow included removal of non-informative variables, transformation of skewed variables, encoding of categorical variables, cyclical encoding of temporal variables, correlation-based feature reduction, and feature scaling.

The variables record ID, date, and country code were removed because they were either non-informative or redundant with other variables. Several positively skewed variables, including vector disease risk score, heat-related admissions, and GDP per capita, were transformed using $\log(x + 1)$. Categorical variables were encoded according to their measurement type: income level was ordinally encoded, while region and country name were one-hot encoded. To represent seasonality, month and week were transformed using sine and cosine functions.

Correlation-based feature reduction was used to remove highly redundant predictors. Feature pairs with absolute correlation greater than 0.85 were examined. AQI was removed because it was used to construct the classification target and was strongly correlated with PM2.5. Week-based cyclical variables and income-level encoding were also removed because their information was largely captured by monthly cyclical variables and continuous socioeconomic indicators. For the sensitivity classification analysis, PM2.5 was additionally removed from the feature set.

Continuous numerical predictors were standardized using Z-score normalization. One-hot encoded variables were retained in binary form and were not standardized. Target variables were excluded from all preprocessing transformations involving feature scaling or model input construction.

The study was organized around four analytical components. First, regression models were used to predict respiratory disease rate from environmental, socioeconomic, geographic, and temporal predictors. Second, classification models were used to predict binary air-quality status under two feature settings: a diagnostic setting with PM2.5 included and a more conservative setting with PM2.5 excluded. Third, model interpretation was conducted to identify the variables most strongly associated with fitted predictions. Fourth, descriptive subgroup analysis was performed across income levels and geographic regions to examine whether aggregate performance concealed differences in error or feature-attribution patterns.

This design allowed the study to evaluate predictive performance, model interpretability, target-proxy dependence in the air-quality classification task, and variation in model behavior across socioeconomic and geographic groups. All findings are interpreted within the scope of the analyzed dataset. The overall workflow of the study is summarized in Figure 1.

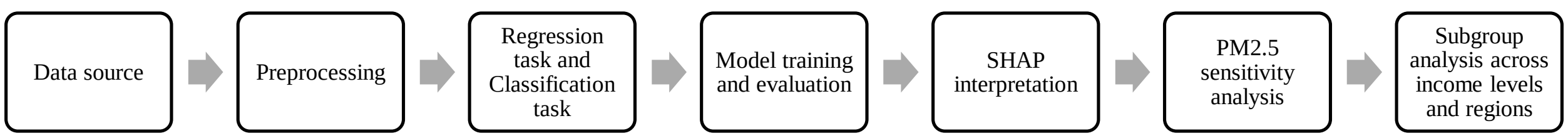

*Figure 1- Overall study workflow*

## 4. Methods

Nine regression models and nine classification models were evaluated. The regression models included Linear Regression, Ridge Regression, Lasso Regression, K-Nearest Neighbors Regressor, Random Forest Regressor, Gradient Boosting Regressor, Support Vector Regression, Multilayer Perceptron Regressor, and XGBoost Regressor. The classification models included Logistic Regression, Decision Tree, K-Nearest Neighbors Classifier, Random Forest Classifier, Gradient Boosting Classifier, Support Vector Machine, Gaussian Naive Bayes, Multilayer Perceptron Classifier, and XGBoost Classifier. These models were selected to represent commonly used linear, nonlinear, ensemble, kernel-based, neural-network, probabilistic, and instance-based approaches for tabular prediction tasks. The models evaluated in each task are summarized in Table 2.

*Table 2- Machine learning models evaluated in this study*

| Model family | Regression models | Classification models |
|---|---|---|
| Linear and regularized models | Linear Regression, Ridge Regression, Lasso Regression | Logistic Regression |
| Tree-based models | — | Decision Tree |
| Ensemble methods | Random Forest Regressor, Gradient Boosting Regressor, XGBoost Regressor | Random Forest Classifier, Gradient Boosting Classifier, XGBoost Classifier |
| Kernel-based models | Support Vector Regression | Support Vector Machine |
| Neural networks | Multilayer Perceptron Regressor | Multilayer Perceptron Classifier |
| Instance-based learning | K-Nearest Neighbors Regressor | K-Nearest Neighbors Classifier |
| Probabilistic models | — | Gaussian Naive Bayes |

All models were evaluated using observation-level nested cross-validation. A 10-fold outer loop was used for performance estimation, and a 5-fold inner loop was used for hyperparameter tuning. Hyperparameters were selected within each outer training fold using grid search and then evaluated on the corresponding held-out outer test fold. Negative RMSE was used for regression model selection, and balanced accuracy was used for classification model selection. Models without major tunable hyperparameters were evaluated directly within the outer loop. The same procedure was applied to the regression task, the diagnostic classification task with PM2.5, and the conservative classification task without PM2.5. Because folds were formed at the record level, observations from the same country and different weeks could occur in both training and test folds. The resulting scores estimate within-dataset performance and do not establish generalization to unseen countries or future time periods. The hyperparameter search spaces and selected configurations are reported in Appendix A.

For regression, RMSE was used as the primary metric, with MAE, MAPE, and $R^2$ reported as supplementary metrics. For classification, balanced accuracy was used as the primary metric, with precision, recall, F1-score, sensitivity, specificity, and confusion matrices reported as supplementary metrics.

Statistical comparisons among the regression models were conducted using the RMSE scores obtained from the 10 outer cross-validation folds. A one-way ANOVA was first used to assess whether mean RMSE differed across the nine models. When the test indicated significant differences, paired t-tests were performed for all 36 model pairs using the matched outer-fold scores. A Bonferroni-adjusted significance level of $\alpha = \frac{0.05}{36} = 0.00139$ was applied to control the family-wise error rate. The same statistical comparison procedure was applied to the diagnostic classification tasks. Balanced-accuracy scores from the 10 matched outer folds were compared across the nine classifiers using a one-way ANOVA, followed by paired t-tests for all 36 model pairs.

SHAP analysis was used to identify the most influential predictors. For regression, SHAP values were computed using the XGBoost regression model. For classification, SHAP values were computed using the XGBoost classifier under both feature settings: with PM2.5 and without PM2.5. Mean absolute SHAP values were used to rank global feature importance. For both tasks, a final XGBoost model was retrained on the full dataset using hyperparameters selected during the inner cross-validation loop, and SHAP values were computed from this model.

SHAP-based feature-selection checks were also conducted by retraining models with reduced feature sets based on SHAP importance rankings and comparing their performance with the full-feature models.

Subgroup analysis was conducted across income levels and geographic regions. For regression, subgroup performance was evaluated using RMSE, MAE, $R^2$, and MAPE. For classification, subgroup performance was evaluated using balanced accuracy, precision, F1-score, sensitivity, and specificity. Group-level SHAP analysis was also used to compare feature contribution patterns across income groups.

Lasso Regression was used for the regression subgroup analysis because it achieved the best overall regression performance. XGBoost was used for the classification subgroup analysis because it showed strong predictive performance and supported SHAP-based interpretation.

# 5. Results

## 5.1 Regression Performance

Table 3 summarizes the regression performance of the nine evaluated models. Linear models achieved the strongest overall performance. Lasso Regression produced the lowest mean RMSE, followed closely by Ridge Regression, Linear Regression, and Support Vector Regression. The differences among these top models were small, indicating that respiratory disease rate was largely explained by linear relationships in the feature space. Ensemble models, including Gradient Boosting, XGBoost, and Random Forest, performed slightly worse than the linear models, while the K-Nearest Neighbors Regressor produced the weakest performance.

*Table 3- Regression model performance*

| Rank | Model | Mean RMSE | Std. RMSE |
|---|---|---|---|
| 1 | Lasso Regression | 9.8795 | 0.1381 |
| 2 | Ridge Regression | 9.8822 | 0.1353 |
| 3 | Linear Regression | 9.8822 | 0.1351 |
| 4 | Support Vector Regression | 9.8867 | 0.1359 |
| 5 | Gradient Boosting Regressor | 9.9071 | 0.1437 |
| 6 | XGBoost Regressor | 9.9255 | 0.138 |
| 7 | Random Forest Regressor | 9.9257 | 0.1495 |
| 8 | Multilayer Perceptron Regressor | 10.003 | 0.1327 |
| 9 | K-Nearest Neighbors Regressor | 11.2045 | 0.2015 |

The one-way ANOVA indicated that regression performance differed across the nine models ($F = 78.05$, $p < 0.001$). Bonferroni-corrected paired comparisons identified statistically significant differences in 22 of the 36 model pairs at the adjusted significance level of $\alpha = 0.00139$. Lasso Regression achieved the largest number of significant pairwise wins, outperforming Random Forest, Gradient Boosting, Multilayer Perceptron, XGBoost, and K-Nearest Neighbors. However, Lasso was not significantly different from Linear Regression, Ridge Regression, or Support Vector Regression. These four models therefore formed a statistically comparable top-performing group. K-Nearest Neighbors was significantly outperformed by all eight other regression models and produced the weakest overall performance. Table 4 shows the Regression Pairwise Comparison Wins.

*Table 4- Regression Pairwise Comparison Wins*

| Rank | Model | Wins |
|---|---|---|
| 1 | Lasso | 5 |
| 2 | Linear Regression | 4 |
| 2 | Ridge | 4 |
| 2 | SVR | 4 |
| 5 | Gradient Boosting | 3 |
| 6 | Random Forest | 2 |
| 6 | XGBoost Regressor | 2 |
| 8 | MLP Regressor | 1 |
| 9 | KNN Regressor | 0 |

## 5.2 Air-Quality Classification Performance

Table 5 summarizes the diagnostic air-quality classification results with PM2.5 included. The top seven classification models achieved closely clustered within-dataset performance, with mean balanced accuracy values between 0.9259 and 0.9285. Gradient Boosting achieved the highest mean balanced accuracy, followed closely by Support Vector Machine, XGBoost, Random Forest, Logistic Regression, Decision Tree, and Multilayer Perceptron. K-Nearest Neighbors and Gaussian Naive Bayes performed substantially worse. Because PM2.5 is closely related to the AQI-derived label, these scores are interpreted primarily as a diagnostic of feature-target dependence.

*Table 5- Classification model performance with PM2.5 included*

| Rank | Model | Mean balanced accuracy | Std. |
|---|---|---|---|
| 1 | Gradient Boosting Classifier | 0.9285 | 0.0076 |
| 2 | Support Vector Machine | 0.928 | 0.0089 |
| 3 | XGBoost Classifier | 0.928 | 0.0074 |
| 4 | Random Forest Classifier | 0.9277 | 0.0072 |
| 5 | Logistic Regression | 0.9274 | 0.0088 |
| 6 | Decision Tree | 0.9268 | 0.0081 |
| 7 | Multilayer Perceptron Classifier | 0.9259 | 0.0073 |
| 8 | K-Nearest Neighbors Classifier | 0.8583 | 0.0105 |
| 9 | Gaussian Naive Bayes | 0.8442 | 0.0076 |

The one-way ANOVA indicated differences in balanced accuracy across the nine classifiers ($F = 152.54$ ,$p < 0.001$. Bonferroni-corrected paired comparisons identified statistically significant differences in 14 of the 36 model pairs at the adjusted significance level of $\alpha = 0.00139$. No statistically significant differences were found among Logistic Regression, Decision Tree, Random Forest, Gradient Boosting, Support Vector Machine, Multilayer Perceptron, and XGBoost. These seven models therefore formed a statistically comparable top-performing group, with mean balanced accuracy ranging from 0.9259 to 0.9285. Each of the seven models significantly outperformed K-Nearest Neighbors and Gaussian Naive Bayes. K-Nearest Neighbors also significantly outperformed Gaussian Naive Bayes.

## 5.3 Sensitivity Analysis Without PM2.5

Because PM2.5 was closely associated to the AQI-derived classification label, a sensitivity analysis removed PM2.5 from the feature set. Specifically, PM2.5 concentration was correlated with the binary air-quality label at $r = 0.810$, quantifying the strength of this target-proxy relationship. Table 6 reports the results under this more conservative setting. Balanced accuracy decreased for all models, confirming that the high performance in the diagnostic setting depended strongly on pollutant information closely related to the target definition.

*Table 6- Classification model performance without PM2.5*

| Rank | Model | Mean balanced accuracy | Std. |
|---|---|---|---|
| 1 | Gradient Boosting Classifier | 0.8172 | 0.0095 |
| 2 | Decision Tree | 0.8171 | 0.0094 |
| 2 | Random Forest Classifier | 0.8171 | 0.0094 |
| 4 | XGBoost Classifier | 0.817 | 0.0091 |
| 5 | Support Vector Machine | 0.8169 | 0.0094 |
| 6 | Multilayer Perceptron Classifier | 0.8163 | 0.0088 |
| 7 | K-Nearest Neighbors Classifier | 0.8083 | 0.0107 |
| 8 | Logistic Regression | 0.8075 | 0.0108 |
| 9 | Gaussian Naive Bayes | 0.7763 | 0.0113 |

The one-way ANOVA indicated differences in balanced accuracy across the nine classifiers in the no-PM2.5 setting ($F=16.55$, $p < 0.001$. Bonferroni-corrected paired comparisons showed no statistically significant differences among Gradient Boosting, Decision Tree, Random Forest, XGBoost, and Support Vector Machine. These five models therefore formed a statistically comparable top-performing group, with mean balanced accuracy ranging from 0.8169 to 0.8172. Logistic Regression, Multilayer Perceptron, and K-Nearest Neighbors each significantly outperformed only Gaussian Naive Bayes, which produced the weakest performance.

## 5.4 SHAP-Based Interpretation

SHAP analysis was used to identify the main predictors contributing to the regression model output. As shown in Figure 2, PM2.5 concentration was by far the dominant predictor of respiratory disease rate, with a mean absolute SHAP value much larger than all other variables. This indicates that variation in predicted respiratory disease rates was strongly associated with PM2.5 concentration in the fitted model. The remaining predictors had comparatively small contributions, including heat-wave days, healthcare access index, extreme weather events, temperature anomaly, precipitation, and vector disease risk score. This pattern is consistent with the regression results, where linear models performed strongly, suggesting that the target was largely influenced by a dominant pollution-related signal. Although socioeconomic and climate-related variables contributed less than PM2.5, their nonzero SHAP values indicate that they still provided secondary explanatory information.

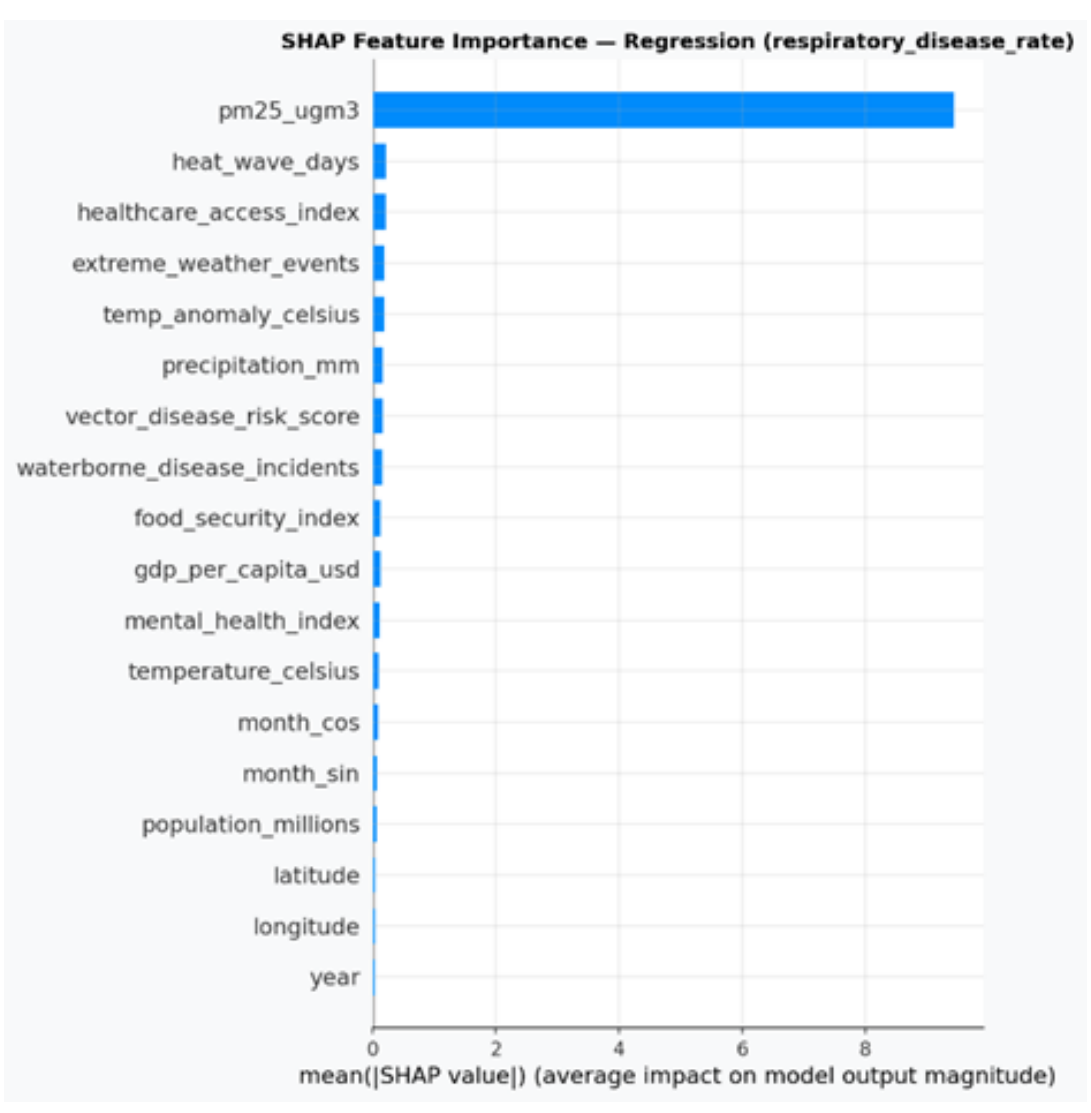

*Figure 2- SHAP feature importance for respiratory disease prediction*

For the diagnostic air-quality classification task, PM2.5 was again the dominant XGBoost predictor, with a mean absolute SHAP value much larger than all other variables. This result illustrates the target-proxy dependence created by predicting an AQI-derived label from a pollutant closely related to AQI. After PM2.5 was removed, the feature-importance pattern changed substantially. GDP per capita became the most influential predictor in the fitted XGBoost model, followed by precipitation and healthcare access index. Other variables, including mental health index, temperature anomaly, and month cosine, contributed smaller amounts. These attributions describe model behavior within the dataset and should not be interpreted as causal determinants of air quality. Figures 3 and 4 compare the SHAP rankings with and without PM2.5.

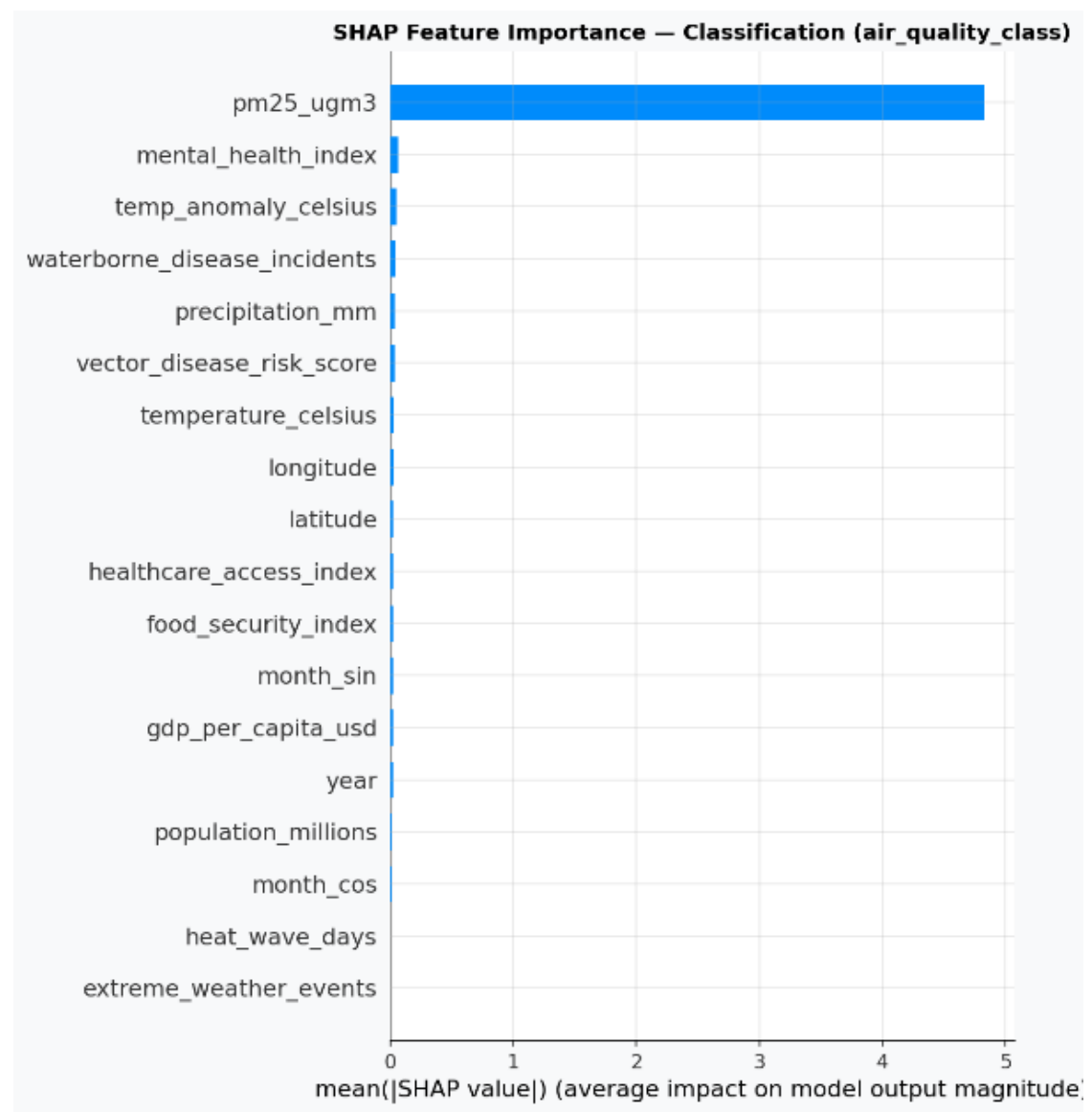

*Figure 3- SHAP feature importance for air-quality classification with PM2.5 included.*

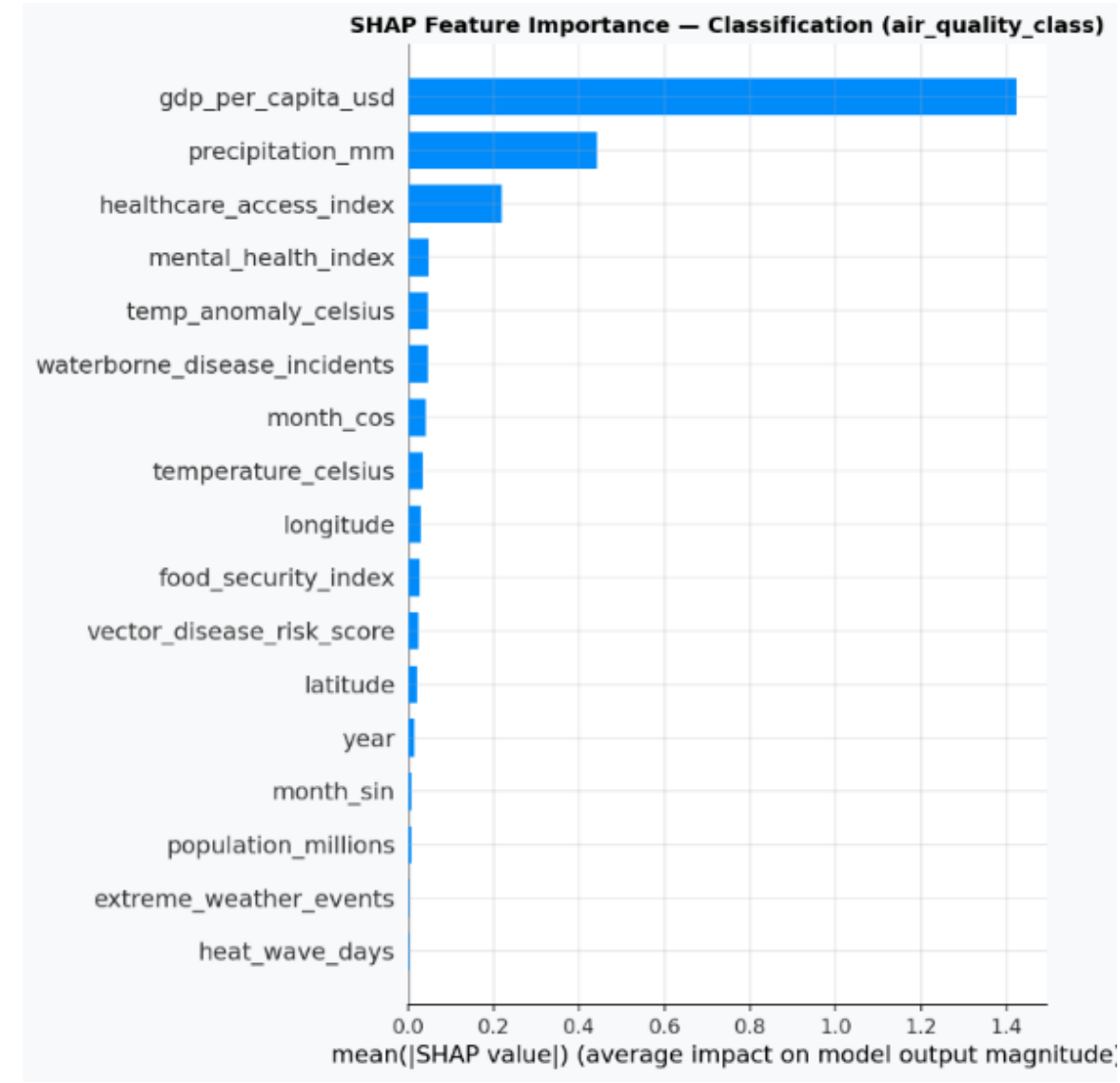

*Figure 4- SHAP feature importance for air-quality classification without PM2.5.*

The SHAP-based feature-selection checks showed that reduced feature sets retained nearly all predictive performance. In the no-PM2.5 classification setting, reducing the feature set from 17 to 12 variables changed balanced accuracy from 0.8171 to 0.8169 and F1-score from 0.7881 to 0.7879. These small differences indicate that several low-ranked predictors contributed little additional predictive information.

## 5.5 Subgroup Analysis

Subgroup analysis was conducted to examine whether model performance and feature contribution patterns differed across income levels and geographic regions. For the regression task, Lasso Regression produced similar aggregate prediction errors across income groups. As shown in Figure 5(a), RMSE values were close across high-income, upper-middle-income, and lower-middle-income countries, with values of 9.86, 9.68, and 10.05, respectively. The RMSE disparity ratio across income groups was 1.038, indicating limited variation in overall prediction error.  Table 7 reports the full set of regression metrics across income groups. MAE and $R^2$ values were similarly consistent across groups, while MAPE was slightly lower for lower-middle-income countries, likely reflecting higher baseline disease rates in that group.

*Table 7- Regression subgroup performance by income level using Lasso*

| Income Level | N | RMSE | MAE | $R^2$ | MAPE |
|---|---|---|---|---|---|
| High | 5,076 | 9.86 | 7.90 | 0.377 | 13.46% |
| Upper-Middle | 3,948 | 9.68 | 7.76 | 0.386 | 13.08% |
| Lower-Middle | 5,076 | 10.05 | 8.02 | 0.380 | 10.28% |

Regional differences were also modest, as shown in Figure 5(b). Most regions had RMSE values close to the overall mean, although Africa showed the highest error and South America showed the lowest error. This suggests that the model maintained relatively consistent regression performance across geographic regions, while still reflecting some regional heterogeneity in the pollution-health relationship.

However, the SHAP-based subgroup analysis revealed a more meaningful difference in feature contribution patterns. As shown in Figure 5(c), PM2.5 was the dominant predictor across all income groups, but its contribution was substantially larger for lower-middle-income countries (mean |SHAP| = 12.47) than for high-income (7.75) and upper-middle-income countries (7.82), representing approximately 60% greater reliance on PM2.5 in lower-income settings. This indicates that while aggregate prediction errors were similar across income groups, the model relied more heavily on PM2.5 when predicting respiratory disease rates in lower-middle-income settings. These results suggest that similar overall model performance can mask important differences in the predictors driving model output across socioeconomic groups.

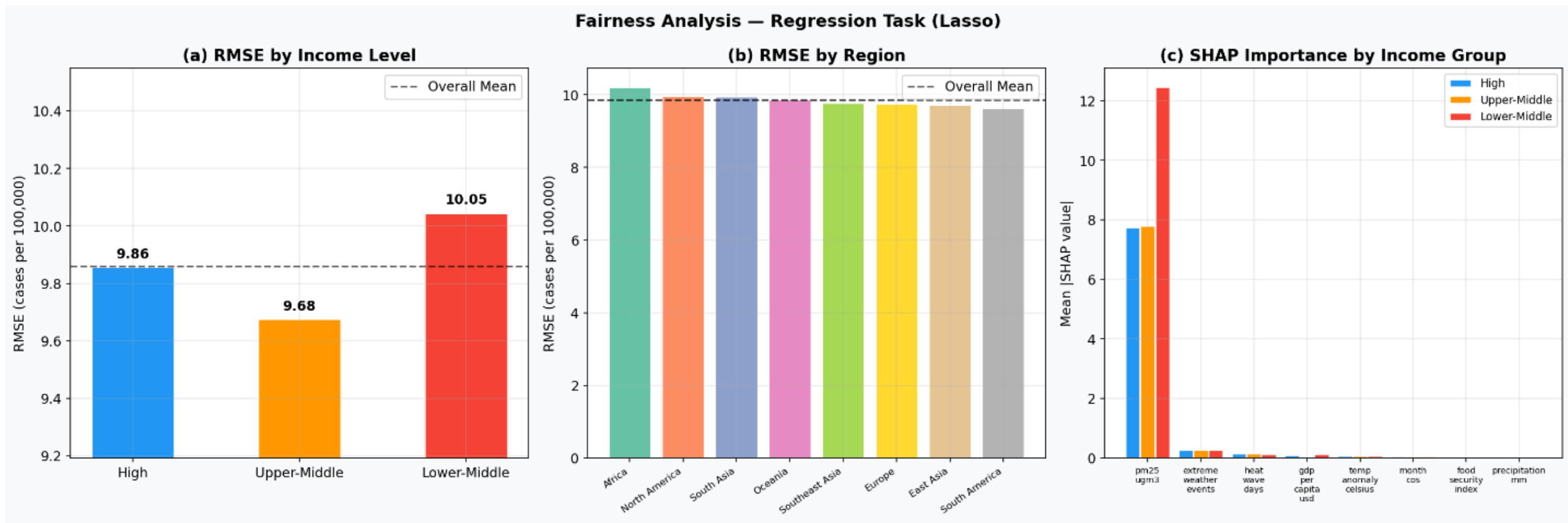

*Figure 5- Regression subgroup analysis using the Lasso model. (a) RMSE by income level. (b) RMSE by geographic region. (c) SHAP feature importance by income group.*

For the classification task, subgroup analysis was conducted using the XGBoost classifier. As shown in Table 8, balanced accuracy was relatively similar across income groups, with values of 0.8765 for high-income countries, 0.8661 for upper-middle-income countries, and 0.8557 for lower-middle-income countries. The balanced accuracy gap across income groups was 0.021, suggesting limited disparity in overall classification performance.

*Table 8- Classification subgroup performance by income level using XGBoost*

| Income level | N | Balanced accuracy | Precision | F1-score | Sensitivity | Specificity |
|---|---|---|---|---|---|---|
| High | 5,076 | 0.8765 | 0.838 | 0.8158 | 0.7946 | 0.9584 |
| Upper-Middle | 3,948 | 0.8661 | 0.841 | 0.8045 | 0.7711 | 0.9612 |
| Lower-Middle | 5,076 | 0.8557 | 0.9684 | 0.9765 | 0.9848 | 0.7266 |

However, Table 8 also shows a more pronounced difference in class-specific metrics. High-income countries had lower sensitivity but higher specificity, while lower-middle-income countries showed the opposite pattern. Specifically, sensitivity increased from 0.7946 in high-income countries to 0.9848 in lower-middle-income countries, whereas specificity decreased from 0.9584 to 0.7266. This indicates that the classifier was more effective at identifying unhealthy air-quality records in lower-middle-income countries, while it was more effective at identifying good air-quality records in high-income countries. This asymmetry likely reflects differences in class distribution and pollution profiles across income groups.

At the regional level, Southeast Asia and Africa showed the highest balanced accuracy values, while South Asia showed the lowest balanced accuracy. These regional differences suggest that model performance varied across geographic contexts, even when aggregate performance appeared strong.

As shown in Figure 6, group-level SHAP analysis further indicated that PM2.5 remained the dominant predictor across all income groups. Its mean absolute SHAP contribution was higher in lower-middle-income countries than in high-income and upper-middle-income countries, suggesting that PM2.5 played a stronger role in air-quality classification for lower-income settings. Overall, these results show that aggregate classification performance can mask meaningful differences in sensitivity, specificity, and feature contribution patterns across socioeconomic groups.

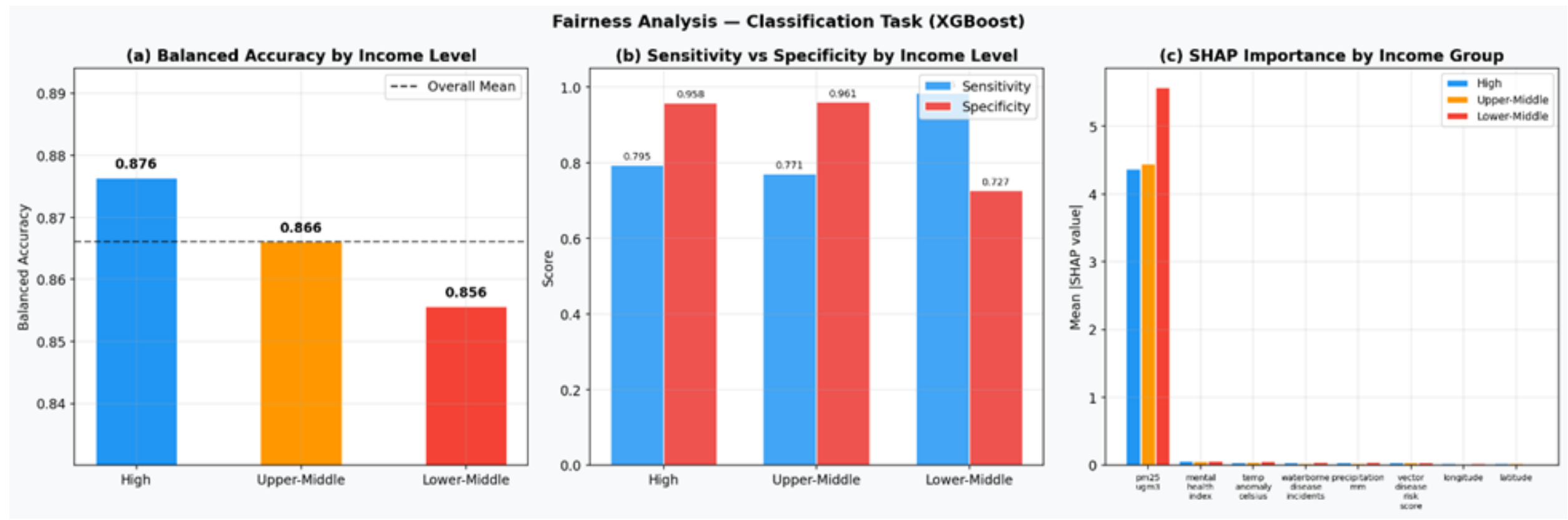

*Figure 6- Classification subgroup analysis using the XGBoost model. The figure compares classification performance and SHAP feature contribution patterns across income groups.*

## 6. Discussion

The results show that PM2.5 plays a central role in predicting respiratory disease rate in this dataset. Across the regression models, PM2.5 was the strongest predictor by a large margin, and the best-performing models were mainly linear or regularized linear models. This suggests that the relationship between air pollution and respiratory disease rate in this dataset is largely driven by a strong and relatively direct pollution-related signal. In other words, adding more complex nonlinear models did not substantially improve prediction because much of the explainable variation was already captured by PM2.5 and a small set of related variables.

The dominance of PM2.5 is still methodologically informative because it reveals that model performance depends heavily on one feature. Such dependence should be identified before using a model for scientific interpretation or decision support. In a future study using traceable monitoring and health-surveillance data, the same workflow could test whether PM2.5 remains dominant after country-grouped and temporal validation and after controlling for measured confounding factors. The smaller contributions of heat-wave days, healthcare access, extreme weather events, temperature anomaly, and precipitation should likewise be treated as hypotheses for further study rather than policy effects.

The classification analysis provides a clear example of target-proxy dependence. When PM2.5 was included, most models achieved high and similar balanced accuracy because PM2.5 is closely related to the AQI used to construct the target. After PM2.5 was removed, performance decreased and the XGBoost model relied more on GDP per capita, precipitation, and healthcare access. These variables may encode indirect structure in the dataset, including country and regional differences, but their SHAP values do not show that they independently determine air quality. The no-PM2.5 setting is therefore the more informative test of whether the remaining variables contain predictive signal beyond the dominant pollutant proxy.

The subgroup analysis further illustrates the difference between aggregate error and model attribution. Regression RMSE was similar across income groups, yet XGBoost SHAP values showed stronger reliance on PM2.5 for lower-middle-income records. This is a difference in fitted-model behavior, not direct evidence of unequal biological effects or environmental burden. It could arise from predictor distributions, country composition, label construction, or the dataset construction process. Confirming a socioeconomic difference would require traceable exposure and health data, appropriate confounder control, and validation that holds out entire countries or regions.

The classification subgroup results showed a related asymmetry. Lower-middle-income records had higher sensitivity for the unhealthy class, whereas high-income records had higher specificity for the good class. The practical meaning of these errors would depend on the intended use of the model, but the present analysis does not evaluate deployment. The result is best viewed as a diagnostic showing why balanced accuracy alone may conceal class-specific differences across subgroups.

Overall, this study demonstrates how interpretable machine learning can be used to examine dominant predictive signals, target-proxy dependence, and subgroup-specific model behavior within a structured public dataset. The findings should be validated using independently traceable environmental and health data before being used for epidemiological interpretation or decision support.

## 7. Limitations and Future Work

This study has several limitations. First, the analysis used a publicly available structured dataset designed to represent climate-health relationships across countries and time. Therefore, the findings should be interpreted as evidence of the proposed analytical workflow rather than as direct epidemiological estimates. Future work should validate the framework using real-world air-quality monitoring data, health surveillance records, and environmental exposure datasets.

Second, although nested cross-validation was used to reduce bias in model selection, the evaluation did not fully account for country-level clustering or temporal dependence. Future studies should apply grouped, leave-one-country-out, and time-based validation to better assess generalization across countries and future periods. Additionally, the paired t-tests used for pairwise model comparison assume independence among cross-validation fold scores, an assumption that is technically violated in K-fold cross-validation. Corrections such as the Nadeau-Bengio adjusted t-test would provide more conservative significance estimates, and the reported significance results should be interpreted with this caveat in mind.

Third, the air-quality classification target was derived from AQI. Although AQI was removed from the feature set and a sensitivity analysis was conducted without PM2.5, the classification results should be interpreted with this target construction in mind. Since AQI thresholds are partially constructed from PM2.5 measurements, the classification task may inherently favor pollutant-based predictors regardless of feature selection decisions. This feature-target dependence cannot be fully addressed without adopting health-outcome-based classification targets. Future work should consider more health-centered classification outcomes, such as high respiratory hospitalization risk or public health alert categories.

Finally, the subgroup analysis was limited to income level and region. Future work should include additional vulnerability indicators, such as age structure, urbanization, baseline disease burden, healthcare capacity, and environmental regulation measures. Extending the framework with real-world data and stronger spatial-temporal validation would provide more reliable evidence for environmental health decision-making.

## 8. Conclusion

This study developed and evaluated an interpretable machine learning framework for predicting climate- and pollution-related health outcomes using country-level weekly data. The analysis combined regression, classification, SHAP-based interpretation, sensitivity analysis, and subgroup evaluation to examine not only model accuracy, but also the environmental and socioeconomic meaning of model behavior.

The results showed that PM2.5 was the dominant predictor of respiratory disease rate. Linear and regularized linear models performed strongly in the regression task, suggesting that predicted respiratory disease burden in this dataset was largely driven by a strong pollution-related signal. For air-quality classification, most models achieved high balanced accuracy when PM2.5 was included. However, performance decreased when PM2.5 was removed, indicating that the classification task depended heavily on pollutant-related information. In the absence of PM2.5, socioeconomic and meteorological variables such as GDP per capita, precipitation, and healthcare access became more important predictors.

The SHAP and subgroup analyses provided additional insight beyond aggregate performance metrics. PM2.5 remained the most influential variable overall, but its contribution was stronger in lower-middle-income countries. The classification subgroup results also showed that similar balanced accuracy across income groups can still hide differences in sensitivity and specificity. These findings highlight the importance of evaluating environmental health models across socioeconomic and geographic groups, especially when the models may be used to support public health surveillance, air-quality alerts, or resource allocation.

Overall, the study suggests that interpretable machine learning can support environmental health analysis by identifying key pollution-related predictors, testing the robustness of prediction tasks, and revealing subgroup differences in model behavior. While the findings should be validated using real-world air-quality monitoring and health surveillance data, the proposed framework provides a useful foundation for transparent, policy-relevant, and equity-aware climate-health prediction.

## Data Availability Statement

The dataset used in this study is publicly available as the Global Climate-Health Impact Tracker dataset through Kaggle (Gokhale 2025). Processed data and analysis code can be made available upon reasonable request.

## Funding

This research received no external funding.

## Conflict of Interest

The authors declare no conflict of interest.

## Author Contributions

The authors were responsible for the study design, data preprocessing, model development, statistical analysis, interpretation of results, and manuscript preparation.

## Appendix A. Hyperparameter Search Spaces

| Model | Task | Parameters | Search Space | Best Value |
|---|---|---|---|---|
| Linear Regression | Regression | — | — | — |
| Ridge Regression | Regression | alpha | [0.1, 1, 10, 100] | 10 |
| Lasso Regression | Regression | alpha | [0.001, 0.01, 0.1, 1] | 0.1 |
| Random Forest | Regression | n_estimators, max_depth | [100], [5, 10, None] | 100, 5 |
| Gradient Boosting | Regression | n_estimators, max_depth, learning_rate | [100,200], [3,5], [0.05,0.1] | 100, 3, 0.05 |
| SVR | Regression | C, kernel | [0.1,1,10,100], [rbf, linear] | 1, linear |
| MLP Regressor | Regression | hidden_layer_sizes, activation | [(64,),(128,),(64,32)],[relu, tanh] | (128,), tanh |
| XGBoost Regressor | Regression | n_estimators, max_depth, learning_rate, subsample | [100], [3,5],[0.1],[0.8] | 100, 3, 0.1, 0.8 |
| KNN Regressor | Regression | n_neighbors | [3, 5, 7, 11] | 11 |
| Logistic Regression | Classification | C | [0.01, 0.1, 1, 10] | 1 |
| Decision Tree | Classification | max_depth, min_samples_split | [3,5,10, None], [2,5,10] | 3, 2 |
| Random Forest | Classification | n_estimators, max_depth | [100], [5,10, None] | 100, 10 |
| Gradient Boosting | Classification | n_estimators, max_depth, learning_rate | [100,200], [3,5], [0.05,0.1] | 200, 3, 0.05 |
| SVM | Classification | C, kernel | [0.01,0.1,1,10,100], [linear, rbf] | 100, linear |
| Naive Bayes | Classification | — | — | — |
| MLP Classifier | Classification | hidden_layer_sizes, activation | [(64,),(128,),(64,32)],[relu, tanh] | (64,), tanh |
| XGBoost Classifier | Classification | n_estimators, max_depth, learning_rate, subsample | [100],[3,5],[0.1],[0.8] | 100, 3, 0.1, 0.8 |
| KNN Classifier | Classification | n_neighbors | [3, 5, 7, 11] | 7 |